\documentclass[aos,preprint]{imsart}
\pdfoutput=1

\usepackage[OT1]{fontenc}
\usepackage[round]{natbib}
\usepackage[colorlinks,citecolor=blue,urlcolor=blue]{hyperref}
\usepackage{hypernat}
\usepackage{amsmath}
\usepackage{amssymb}
\usepackage{amsthm}
\usepackage{graphicx}
\usepackage{caption}
\usepackage{subfig}
\usepackage{float}
\usepackage{placeins}
\startlocaldefs
\usepackage{macros}
\endlocaldefs

\setattribute{journal}{name}{}

\usepackage[
  top=2.5cm,
  bottom=2.5cm,
  left=3cm,
  right=3cm
]{geometry}

\begin{document}
\begin{frontmatter}
  \title{ClusterCluster: Parallel Markov Chain Monte Carlo for Dirichlet Process Mixtures}

  \runtitle{ClusterCluster: Parallel Markov Chain Monte Carlo}                                                               

  \begin{aug}

    \author{\fnms{Dan} \snm{Lovell}%
      \ead[label=e1]{dlovell@alum.mit.edu}},
    \author{\fnms{Jonathan} \snm{Malmaud}%
      \ead[label=e2]{malmaud@mit.edu}},
    \author{\fnms{Ryan P.} \snm{Adams}%
      \ead[label=e3]{rpa@seas.harvard.edu}}%
    \and
    \author{\fnms{Vikash K.} \snm{Mansinghka}%
      \ead[label=e4]{vkm@mit.edu}}%

    \address{Dan Lovell\\      
      Massachusetts Institute of Technology\\
      \printead{e1}}

    \address{Jonathan M. Malmaud\\      
      Massachusetts Institute of Technology\\
      \printead{e2}}

    \address{Ryan P. Adams\\
      School of Engineering and Applied Sciences\\
      Harvard University\\
      \printead{e3}}

    \address{Vikash K. Mansinghka\\
      Computer Science and Artificial Intelligence Laboratory\\
      Massachusetts Institute of Technology\\
      \printead{e4}}

    \affiliation{Massachusetts Institute of Technology and Harvard University}

    \runauthor{Lovell, Malmaud, Adams, Mansinghka}
  \end{aug}

\begin{abstract}
The Dirichlet process (DP) is a fundamental mathematical tool for Bayesian nonparametric modeling, and is widely used in tasks such as density estimation, natural language processing, and time series modeling. Although MCMC inference methods for the DP often provide a gold standard in terms asymptotic accuracy, they can be computationally expensive and are not obviously parallelizable. We propose a reparameterization of the Dirichlet process that induces conditional independencies between the atoms that form the random measure. This conditional independence enables many of the Markov chain transition operators for DP inference to be simulated in parallel across multiple cores. Applied to mixture modeling, our approach enables the Dirichlet process to simultaneously learn clusters that describe the data and superclusters that define the granularity of parallelization. Unlike previous approaches, our technique does not require alteration of the model and leaves the true posterior distribution invariant. It also naturally lends itself to a distributed software implementation in terms of Map-Reduce, which we test in cluster configurations of over 50 machines and 100 cores. We present experiments exploring the parallel efficiency and convergence properties of our approach on both synthetic and real-world data, including runs on 1MM data vectors in 256 dimensions.
\end{abstract}
\end{frontmatter}

\section{Introduction}
Bayesian nonparametric models are a remarkable class of stochastic
objects that enable one to define infinite dimensional random
variables that have tractable finite dimensional projections.  This
projective property often makes it possible to construct probabilistic
models which can automatically balance simplicity and complexity in
the posterior distribution.  The Gaussian process (see, e.g.,
\citet{adler-taylor-2007a, rasmussen-williams-2006a}), Dirichlet
process \citep{ferguson-1973a,ferguson-1974a} and Indian buffet
process \citep{griffiths-ghahramani-2006a,ghahramani-etal-2007a} are
the most common building blocks for Bayesian nonparametric models, and
they have found uses in a wide variety of domains: natural language
models \citep{teh-etal-2006a}, computer vision
\citep{sudderth-etal-2006a}, activity modeling \citep{fox-etal-2009a},
among many others.

Most commonly, Bayesian nonparametric models use the infinite
dimensional construction to place priors on the \emph{latent}
parameters of the model, such as in Dirichlet process mixtures
\citep{escobar-west-1995a,rasmussen-2000a}, Gaussian Cox processes
\citep{moller-etal-1998a,adams-murray-mackay-2009c}, and latent
feature models \citep{fox-etal-2009b}.  This approach to priors for
latent structure is appealing as the evidence for, e.g., a particular
number of components in a mixture, is often weak and we wish to be
maximally flexible in our specification of the model.  Unfortunately,
the use of Bayesian nonparametric priors for latent structure often
yields models whose posterior distribution cannot be directly
manipulated; indeed a density is often unavailable.  In practice, it
is therefore common to perform approximate inference using Markov
chain Monte Carlo (MCMC), in which posterior computations are
performed via Monte Carlo estimates from samples.  These samples are
obtained via a Markov chain that leaves the posterior
distribution invariant.  Remarkably, MCMC moves can be simulated on
practical finite computers for many Bayesian nonparametric models,
despite being infinite-dimensional, e.g., \citet{rasmussen-2000a,
  neal-2000a, walker-2007a, papaspiliopoulos-roberts-2008a,
  adams-murray-mackay-2009a}.  This property arises when
finite data sets recruit only a finite projection of the underlying
infinite object. Most practical Bayesian nonparametric
models of interest are designed with this requirement in mind.

Markov chain Monte Carlo, however, brings with it frustrations.  Chief
among these is the perception that MCMC is computationally expensive
and not scalable.  This is conflated with the observation that the
Markovian nature of such inference techniques necessarily require the
computations to be sequential.  In this paper, we challenge both of
these conventional wisdoms for one of the most important classes of
Bayesian nonparametric model, the Dirichlet process mixture (DPM).  We
take a novel approach to this problem that exploits invariance
properties of the Dirichlet process to reparameterize the random
measure in such a way that conditional independence is introduced
between sets of atoms.  These induced independencies, which are
themselves inferred as part of the MCMC procedure, enable transition
operators on different parts of the posterior to be
simulated in parallel on different hardware, with minimal
communication.  Unlike previous parallelizing schemes such as
\citet{asuncion-etal-2008a}, our approach does not alter the prior
\emph{or} require an approximating target distribution.  We find that
this parallelism results in real-world gains as measured by several
different metrics against wall-clock time.

\begin{figure}[tb]
\centering%
\subfloat[Supercluster 1]{%
  \includegraphics[width=0.23\textwidth]{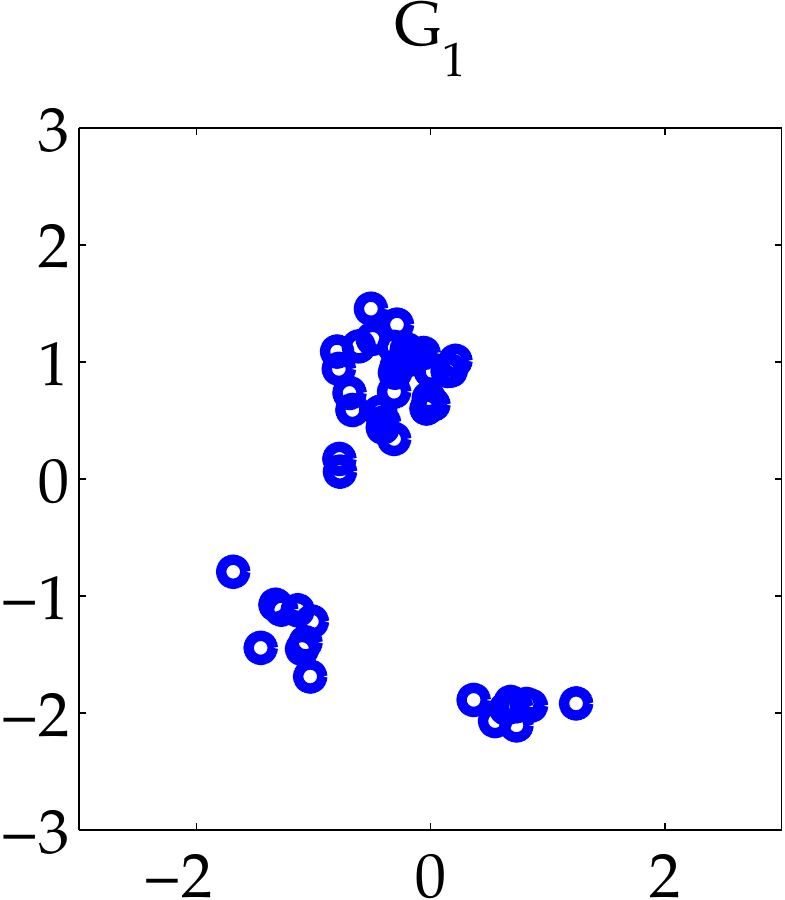}
}~
\subfloat[Supercluster 2]{%
  \includegraphics[width=0.23\textwidth]{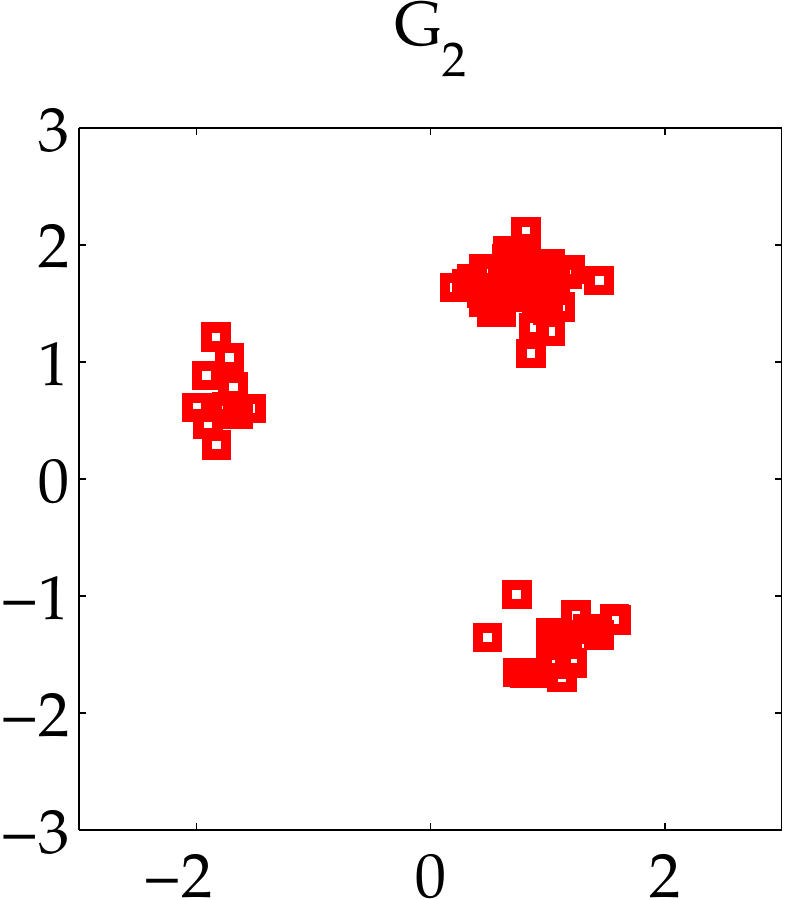}
}~
\subfloat[Supercluster 3]{%
  \includegraphics[width=0.23\textwidth]{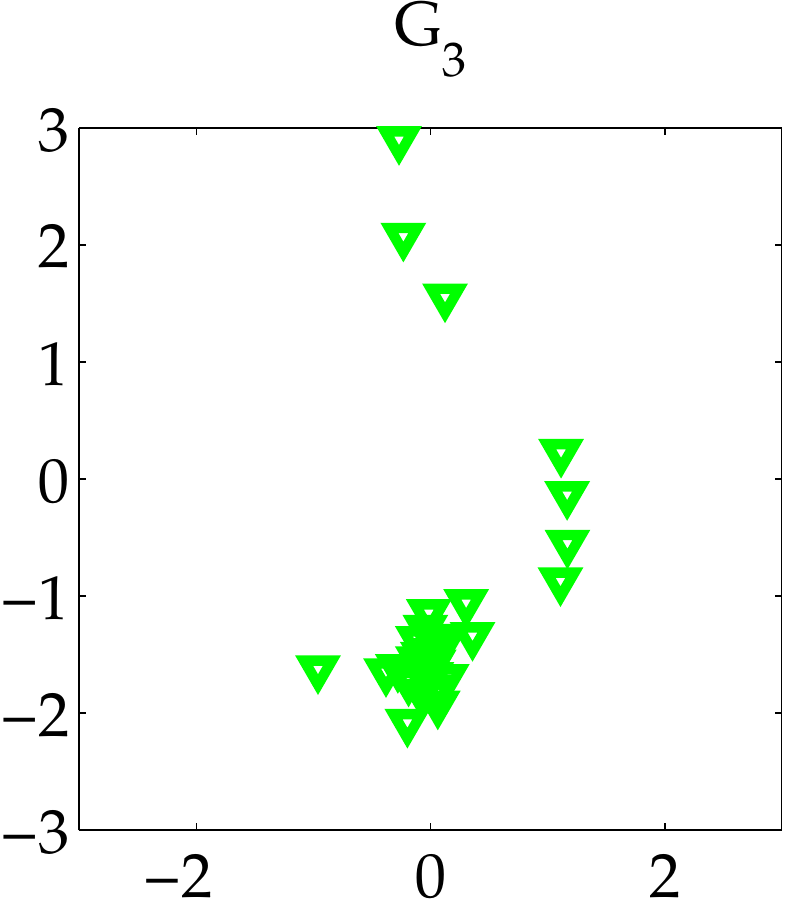}
}~
\subfloat[Supercluster Mixture]{%
  \includegraphics[width=0.23\textwidth]{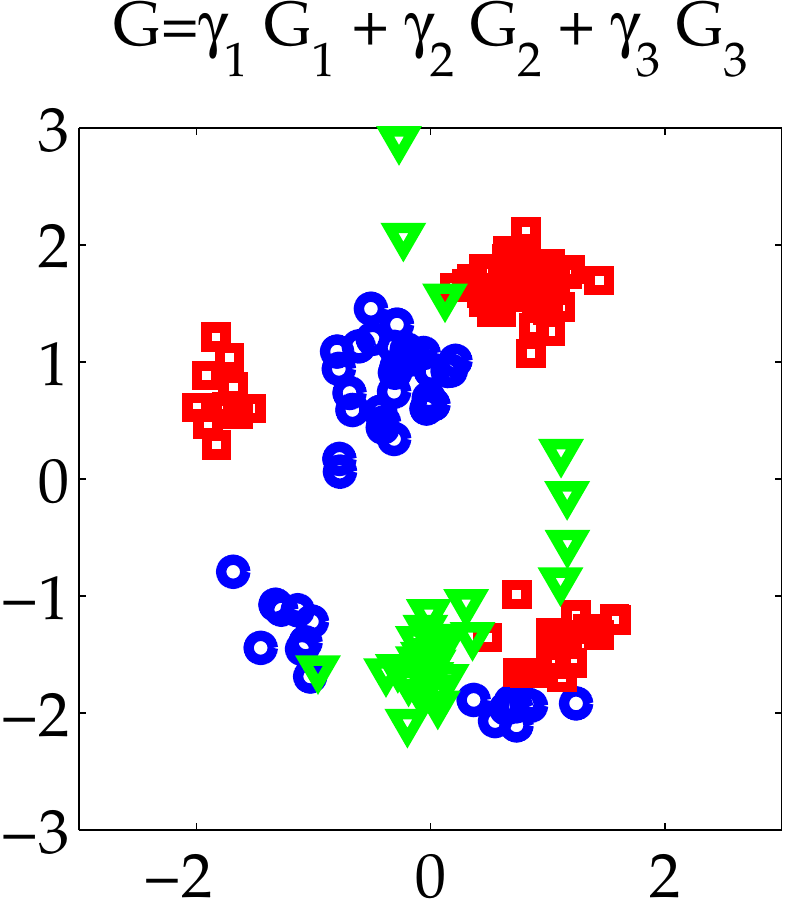}
}
\caption{{\bf Illustration of our auxiliary variable representation
    applied to Dirichlet process mixtures.} (a) to (c) show three
  superclusters, one per compute node, with independent Dirichlet
  process mixtures. (d) shows their linear combination, another
  Dirichlet process mixture. Note that the clusters within each
  supercluster need {\em not} be similar for our scheme to deliver
  efficiency gains.}
\end{figure}

\section{The Dirichlet Process}

The Dirichlet process defines a distribution over probability measures
in terms of a base probability measure~$H$ on a sample space~$\Theta$
and a concentration parameter~${\alpha>0}$.  In its most general form,
a Dirichlet process is characterized by the property that any finite
measurable partition of a~$\Theta$ leads to a finite Dirichlet
distribution over the associated probability measure.  That is,
if~$\Theta$ is partitioned into~${A_1, A_2, \cdots, A_M}$, then the
probability measure~${G(A_1), G(A_2), \cdots, G(A_M)}$ has a finite
Dirichlet distribution with parameters~${\alpha H(A_1), \alpha H(A_2),
  \cdots, \alpha H(A_M)}$.  As an alternative to this somewhat
abstract definition, DP probability measures can also be constructed
from a \emph{stick breaking} process:
\begin{align*}
  G &= \sum^\infty_{j=1}\pi_j \,\delta_{\theta_j} &
  \nu_j\given\alpha &\sim \distBeta(1, \alpha)
\end{align*}
\vspace{-0.5cm}
\begin{align*}
  \theta_j\given H &\sim H &
  \pi_1 &= \nu_1 &
  \pi_j &= \nu_j\prod^{j-1}_{j'=1}(1-\nu_{j'}),
\end{align*}
where it is clear from this construction that the~$G$ are discrete
with probability one.  To achieve continuous density functions, the DP
is often used as a part of an infinite mixture model:
\begin{align}
  F &= \sum^{\infty}_{j=1}\pi_j\,F_{\theta_j},
\end{align}
where~$F_\theta$ is a parametric family of component distributions and
the base measure~$H$ is now interpreted as a prior on this family.
This Dirichlet process mixture model (DPM) is frequently used for
model-based clustering, in which data belonging to a single~$F_\theta$
are considered to form a group.  The Dirichlet process allows for this
model to possess an unbounded number of such clusters.  This view
leads to a related object called the \emph{Chinese restaurant process}
in which the~$\pi_j$ are integrated out and one considers the
infinitely exchangeable distribution over groupings alone.

\section{Nesting Partitions in the Dirichlet Process}
Our objective in this work is to construct an auxiliary-variable
representation of the Dirichlet process in which 1)~the clusters are
partitioned into ``superclusters'' that can be separately assigned to
independent compute nodes; 2)~most Markov chain Monte Carlo transition
operators for DPM inference can be performed in parallel on these
nodes; and 3)~the original Dirichlet process prior is kept intact,
regardless of the distributed representation.  We will assume that
there are~$K$ superclusters, indexed by~$k$.  We will
use~${j\in\naturals}$ to index the clusters uniquely across
superclusters, with~${s_j\in\{1,2,\cdots,K\}}$ being the supercluster
to which~$j$ is assigned.

The main theoretical insight that we use to construct our auxiliary
representation is that the marginal distribution over the mass
allocation of the superclusters arises directly from Ferguson's
definition of the Dirichlet process.  That is, we can generate a
random~$\distDP(\alpha,H)$ partitioning of~$\Theta$ in stages.  First,
choose vector~$\bmu$ on the~$K$-dimensional simplex, i.e., ~$\mu_k\geq
0$ and~${\sum_k\mu_k = 1}$.  Next, draw another vector~$\bgamma$, also
on the~$K$-simplex, from a Dirichlet distribution with base
measure~$\alpha\bmu$:
\begin{align}
\gamma_1, \gamma_2, \cdots, \gamma_K &\sim
\distDir(\alpha\mu_1, \alpha\mu_2, \cdots, \alpha\mu_K).
\end{align}
Then draw~$K$ random distributions from~$K$ independent Dirichlet
processes with base measure~$H$ and concentration
parameters~$\alpha\mu_k$.  These are then mixed together with
the~$\gamma_k$:
\begin{align}
  G_k &\sim \distDP(\alpha\mu_k, H) &
  G &= \sum^K_{k=1}\gamma_k G_k.
\end{align}
This procedure results in~${G\sim\distDP(\alpha, H)}$. Note that the
result of this formulation is a Dirichlet process in which the
components have been partitioned into~$K$ superclusters such that each
contains its own ``local'' Dirichlet process.

Marginalizing out the sticks of each local Dirichlet process results
naturally in a Chinese restaurant process with concentration
parameter~$\alpha\mu_k$.  Interestingly, we can also integrate out
the~$\gamma_k$ to construct a two-stage Chinese restaurant variant.
Each ``customer'' first chooses one of the~$K$ ``restaurants'' according to
its popularity:
\begin{align*}
  \Pr(\mbox{datum $n$ chooses supercluster $k$}\given\alpha)
  &
  = \frac{\alpha\mu_k + \sum_{n'=1}^{n-1} \bbI(s_{z_{n'}} = k)}
  {\alpha + n - 1}.
\end{align*}
This corresponds to the predictive distribution of the
Dirichlet-multinomial over superclusters.  In the second stage, the
customer chooses a table~$z_n$ at their chosen restaurant~$k$
according to its popularity among other customers at that restaurant:
\begin{align*}
  \Pr(z_n = \mbox{extant component } j \given \alpha, s_j = k)
  & 
  = \frac{\sum_{n'=1}^{n-1} \bbI(s_{z_{n'}}=k, z_{n'}=j)}
       { \alpha\mu_k + \sum^{n-1}_{n'=1} \bbI(s_{z_{n'}}=k)}
  \\
  \Pr(z_n = \mbox{new component } j \given \alpha, s_j = k)
  & 
  = \frac{\alpha\mu_k}{ \alpha\mu_k + \sum^{n-1}_{n'=1} \bbI(s_{z_{n'}}=k)}.
\end{align*}

\begin{figure}[t]
  \centering
  \subfloat[]{%
  \includegraphics[height=2in]{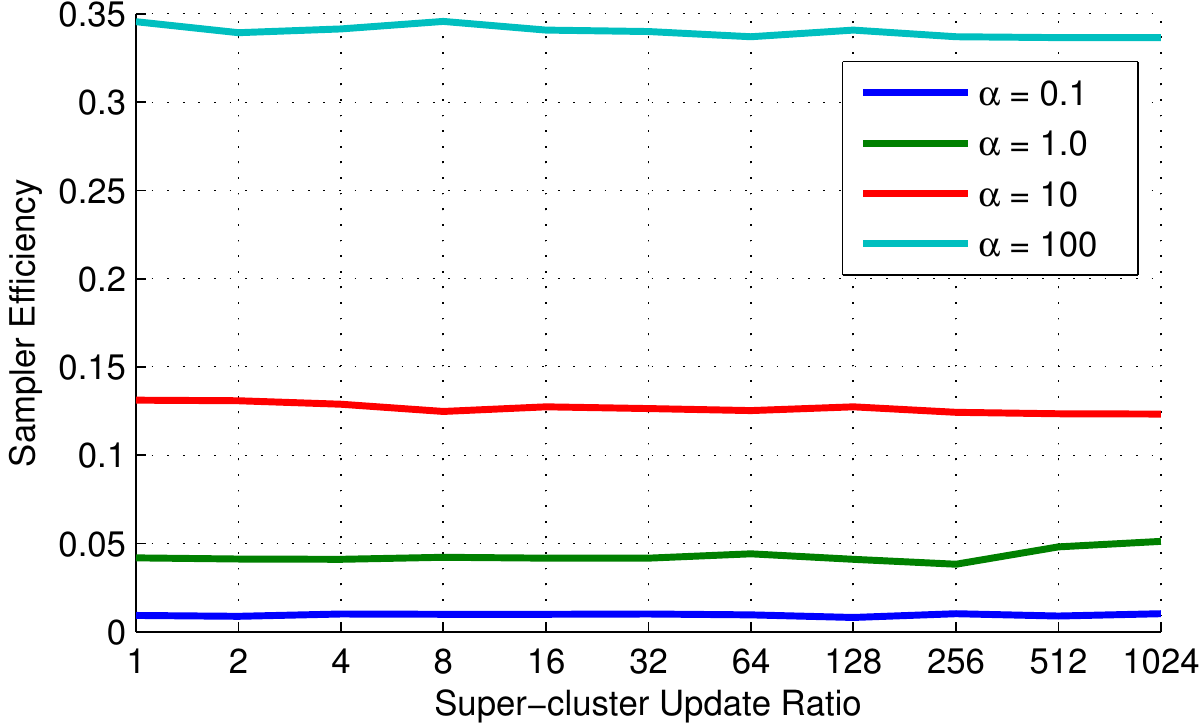}
  }~\hfill%
  \subfloat[]{%
    \includegraphics[height=2in]{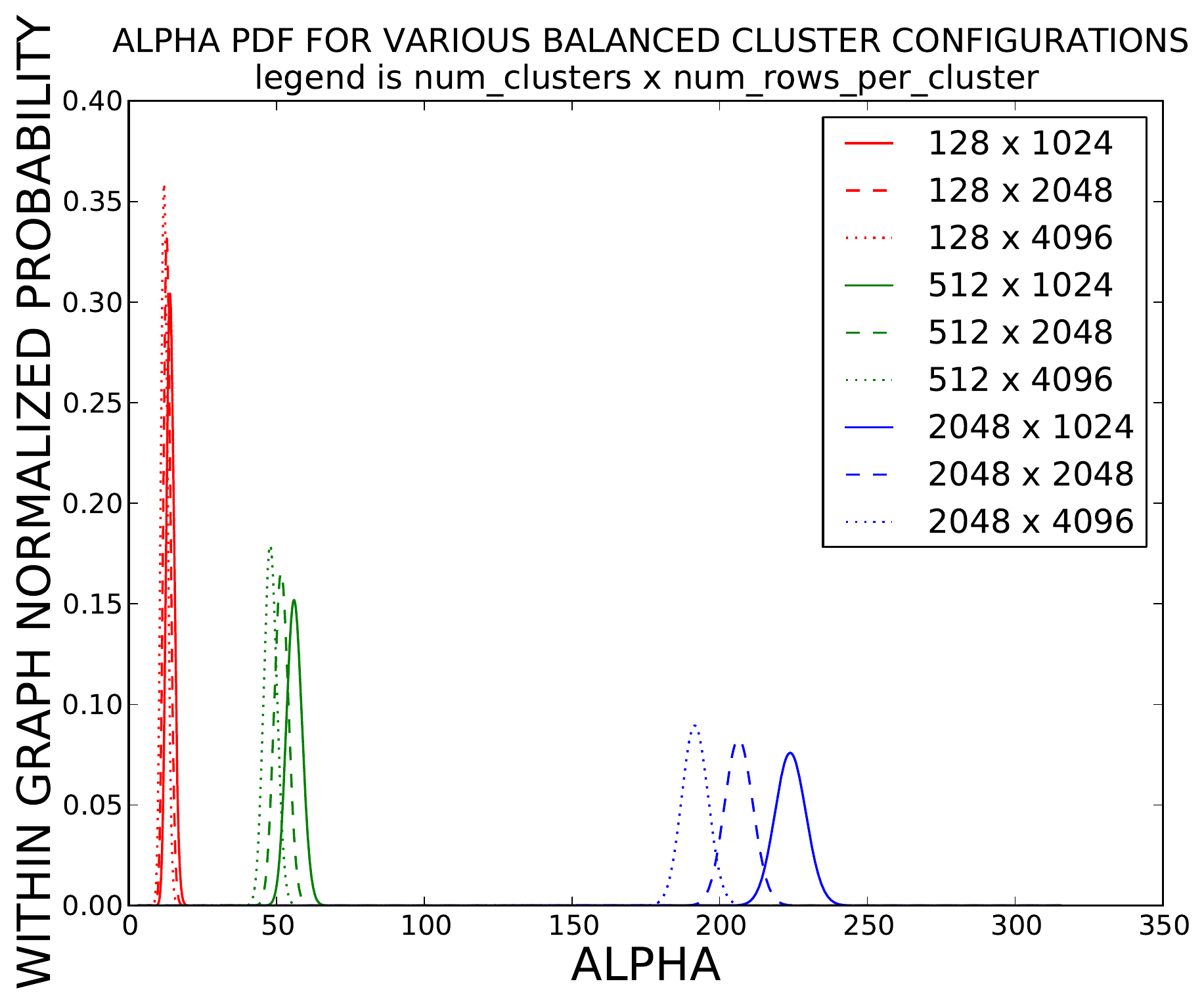}
  }
  \caption{\textbf{Numerical calculations regarding the impact of auxiliary variable scheme on sampler efficiency.}
  (a)Plot of sampling efficiency (effective number of samples
    per MCMC iteration) from the prior as a function of the number of
    local-machine sweeps over the data per cross-machine update.  The
    lines correspond to different concentration parameters.  The
    Chinese restaurant representation was used, with ten
    superclusters, one thousand data and 100,000 iterations. Note that sampler efficiency is roughly independent of super-cluster update ratio and increases with higher concentration parameter, indicating that more parallel gain can be expected the higher the number of latent clusters in the training dataset. 
    (b) The posterior probability distribution on the Dirichlet concentration parameter for various configurations  balanced mixture models. The number of clusters is varied from 128 to 2048, and the number of data points per cluster is varied from 1024 to 4096. As larger concentration parameters imply more room for parallelization in our method,  this view represents the opportunity for parallel gains as a function of the latent structure and quantity of the data.}
\end{figure}

\section{Markov Transition Operators for Parallelized Inference}

\begin{figure*}[t]
\centering
\includegraphics[width=\textwidth]{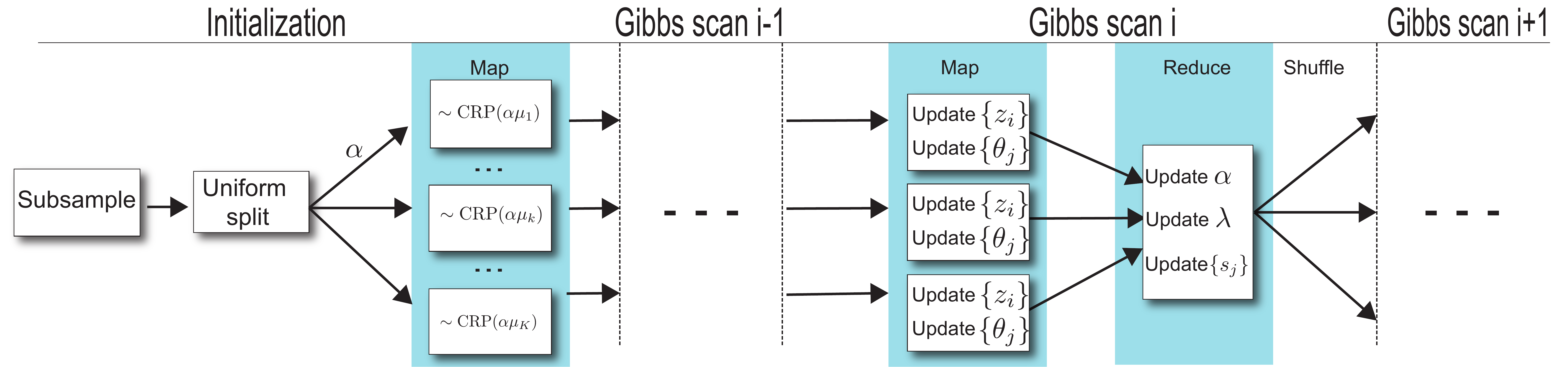}
\caption{{\bf The map-reduce dataflow for our distributed sampler.} 
During initialization, the data are randomly assigned to compute nodes (superclusters). On each MCMC scan, independent mappers perform clustering assuming fixed hyperparameters $\lambda$. These hyperparameters, along with the concentration parameter and assignments of clusters to superclusters are updated in the reduce step. Finally, clusters are shuffled amongst superclusters and the new latent state is communicated to each worker, leaving the cluster in a suitable state for further iterations. The Dirichlet process is effectively learning both how to cluster the data and at what granularity to parallelize inference.}
\label{fig:workflow}
\end{figure*}

The goal of this distributed representation is to provide MCMC
transition operators that can efficiently use multiple cores for
inference.  We have observed~$N$ data~$\{x_n\}^N_{n=1}$ and we wish to
simulate a Markov chain on the associated posterior distribution.
Following the standard approach for mixture modelling, we introduce
latent variables~${z_n\in\naturals}$ that identify the cluster to
which datum~$n$ is assigned.  In the Markov chain, we will also
represent the cluster-specific parameters~$\theta_j$ (although some
model choices allow these to be marginalized away) and the
concentration parameter~$\alpha$.

We introduce some notation for convenience when referring to counts:
\begin{align*}
  \mbox{\small \emph{num data in s.c.}~$k$:\quad} 
  \#_k &= \sum_{n=1}^N \bbI(s_{z_n}=k)\\
  \mbox{\small \emph{num data in cluster}~$j$:\quad}
    \#_j &= \sum_{n=1}^N \bbI(z_n=j)\\
  \mbox{\small \emph{extant clusters in s.c.}~$j$:\quad}
  J_k &= \sum_{j=1}^\infty \bbI(\#_k > 0, s_j=k)
\end{align*}

We use these to examine the prior over the~$z_n$ and $s_j$:
\begin{align}
\Pr(\{z_n\}, \{s_j\} \given \alpha) &=
    \overbrace{
  \left[
  \frac{\Gamma(\alpha)}{\Gamma(N+\alpha)}
  \prod^K_{k=1}
  \frac{\Gamma(\#_k + \alpha\mu_k)}{\Gamma(\alpha\mu_k)}
  \right]
  }^{\mbox{Dirichlet-Multinomial}}
\times
    \overbrace{
  \left[
    \prod^K_{k=1}
    (\alpha\mu_k)^{J_k} \frac{\Gamma(\alpha\mu_k)}{\Gamma(\alpha\mu_k + \#_k)}
    \right]
    }^{\mbox{$K$ independent CRPs}}
    \\
    &=
    \frac{\Gamma(\alpha)}{\Gamma(N+\alpha)}
    \alpha^{\sum^K_{k=1}J_k}
      \prod^K_{k=1}\mu_k^{J_k}.
\end{align}
Note that the terms cancel out so that the result is a marginal
Chinese restaurant process multiplied by a multinomial over how the
components are distributed over the superclusters.

\paragraph{Updating $\alpha$ given the~$z_n$:}
This operation must be centralized, but is lightweight.  Each
supercluster communicates its number of clusters~$J_k$ and these are
used to sample from the conditional (assuming prior~$p(\alpha)$):
\begin{align}
  p(\alpha \given \{z_n\}^N_{n=1}) &\propto
  p(\alpha) \frac{\Gamma(\alpha)}{\Gamma(N+\alpha)}
  \alpha^{\sum_{k=1}^KJ_k}.
\end{align}
This can be done with slice sampling or adaptive rejection sampling.

\paragraph{Updating base measure hyperparameters:}
It is often the case in Bayesian hierarchical models that there are
parameters governing the base measure~$H$.  These are typically
hyperparameters that determine the priors on cluster-specific
parameters~$\theta$ and constrain the behavior of~$F_\theta$.  Updates
to these parameters are performed in the reduce step, based on
sufficient statistics transmitted from the map step.

\paragraph{Updating $\theta_j$ given~$z_n$:}
These are model-specific updates that can be done in parallel, as
each~$\theta_j$ is only asked to explain data that belong to
supercluster~$s_j$. 

\paragraph{Updating $z_n$ given~$s_j$,~$\theta_j$, and~$\alpha$:}
This is typically the most expensive MCMC update for Dirichlet process
mixtures: the hypothesis over cluster assignments must be modified for
each datum.  However, if only \emph{local} components are considered,
then this update can be parallelized.  Moreover, as the
reparameterization induces~$K$ conditonally independent Dirichlet
processes, standard DPM techniques, such as \citet{neal-2000a},
\citet{walker-2007a}, or \citet{papaspiliopoulos-roberts-2008a} can be
used \emph{per supercluster} without modification.  Data cannot move
to components on different machines (in different superclusters), but
can instantiate previously-unseen clusters within its local
superclusters in the standard way.  Note that the~$\mu_k$ scaling
causes these new components to be instantiated with the correct
probability.

\paragraph{Updating $s_j$ given~$z_n$ and $\alpha$:}
As data can only move between clusters that are local to their
machine, i.e., within the same supercluster, it is necessary to move
data between machines.  One efficient strategy for this is to move
entire clusters, along with their associated data to new
superclusters.  This is a centralized update, but it only requires
communicating a set of data indices and one set of component
parameters.  The update itself is straightforward: Gibbs sampling
according to the Dirichlet-multinomial, given the other assignments.
In particular, we note that since~$\theta_j$ moves with the cluster,
the likelihood does not participate in the computation of the
transition operator.  We define~$J_{k\backslash j}$ to be the number
of extant clusters in supercluster~$k$, ignoring cluster~$j$.  The
conditional posterior update is then
\begin{align}
  \Pr(s_j = k \given \{J_{k' \backslash j}\}^K_{k'=1}, \alpha)
  &= \mu_k.
  \frac{\alpha\mu_k + J_{k \backslash j}}
  {\alpha + \sum_{k'=1}^K J_{k'\backslash j}}.
\end{align}

\begin{figure}[t]
\centering
\begin{minipage}[t]{0.49\linewidth}
\centering%
\includegraphics[height=2.6in]{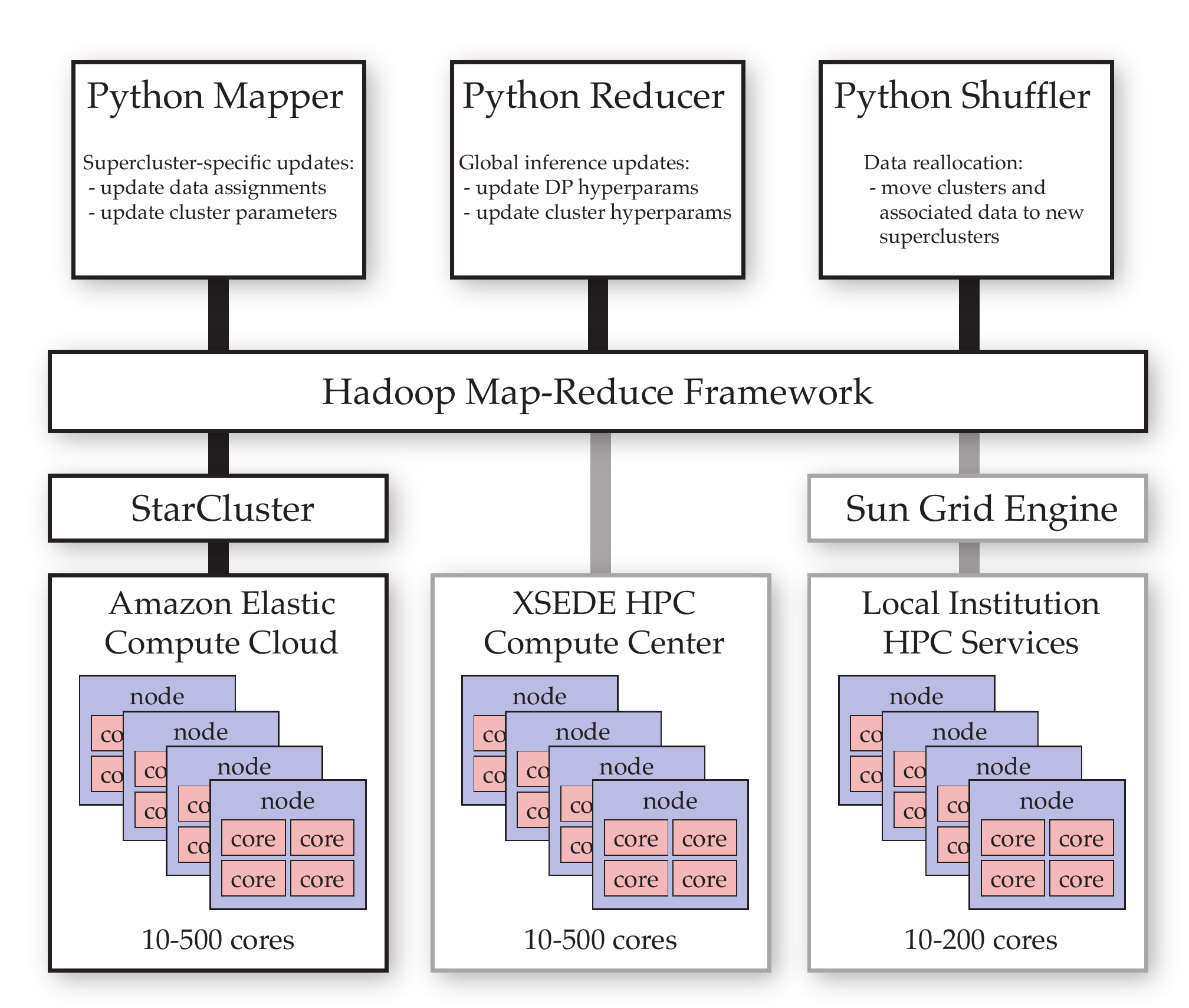}
\caption{\small  {\bf Software architecture for our implementation.} Scaling to a million datapoints was achievable despite only modestly optimized Python implementations of transition operators. Typical runs involved 10 to 50 machines with 2 to 4 cores each. Although we focused on Elastic Compute Cloud experiments, our architecture is appropriate for other distributed computing platforms.}
\label{fig:software_stack}
\end{minipage}
\hfill
\begin{minipage}[t]{0.49\linewidth}
  \centering%
  \includegraphics[height=2.6in]{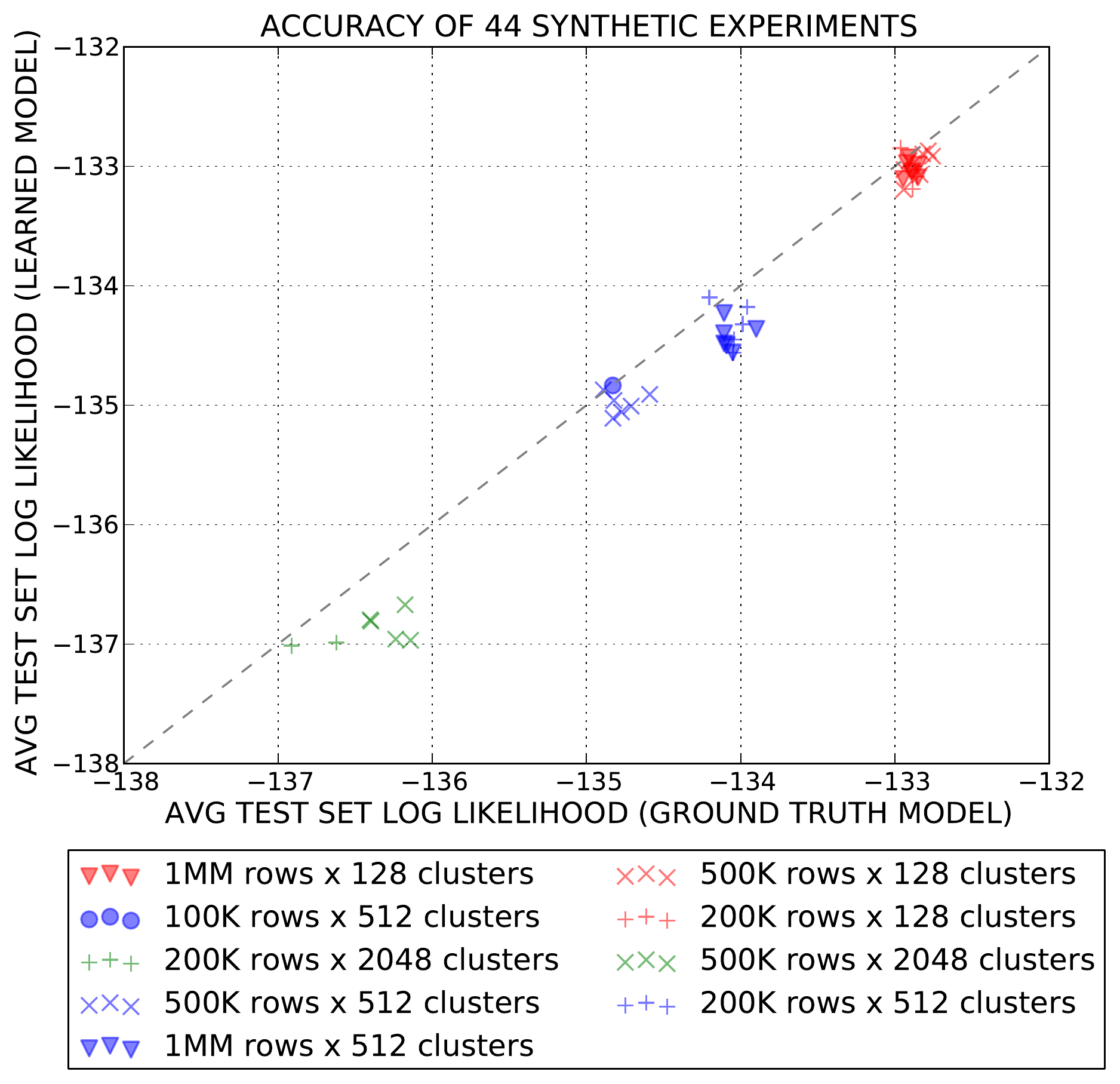}
  \caption{\small \textbf{Our parallel sampler constructs accurate
      density estimates for many synthetic data sources.} We generated
    synthetic datasets from finite mixture models ranging from 200,000
    up to one million datapoints and from 128 clusters up to 2048
    clusters. 
    Marker designates the number of
    clusters, colors indicate the number of datapoints.
    Data is
    jittered for visual clarity.}
  \label{fig:predictive_estimates}
\end{minipage}
\end{figure}

\section{Distributed implementation}

The conditional independencies inherent in the Markov chain transition
operators we have defined correspond naturally with an efficient,
distributed implementation in terms of Map-Reduce \citep{dean2008}.
Figure \ref{fig:workflow} describes the workflow. These operators,
implemented as mappers and reducers, act on a distributed
representation of the latent state, that is also based on the
independencies in our auxiliary variable representation. Intuitively,
each mapper performs MCMC updates on an independent clustering problem
(within the supercluster it corresponds to), assuming fixed
hyperparameters. The reduce step collects the latent state together
and updates hyperparameters, while the shuffle step broadcasts the new
hyperparameters and shuffles clusters amongst the superclusters.

Our system software implementation, described in Figure
\ref{fig:software_stack}, is based on Python implementations, with
modest optimization (in Cython) for the most compute-intensive inner
loops. We use the Hadoop open source framework for Map-Reduce, and
perform experiments using on-demand compute clusters from Amazon's
Elastic Compute Cloud. Typical configurations for our experiments
involved 10-50 machines, each with 2-4 cores, stressing gains due to
parallelism were possible despite significant inter-machine
communication overhead.

We used a uniform prior over the superclusters, i.e.,~${\mu_k=1/K}$.
For initialization, we perform a small calibration run (on 1-10\% of
the data) using a serial implementation of MCMC inference, and use
this to choose the initial concentration parameter~$\alpha$. We then
assign data to superclusters uniformly at random, and initialize the
clustering via a draw from the prior using the local Chinese
restaurant process. This is sufficient to roughly estimate (within an
order of magnitude) the correct number of clusters, which supports
efficient density estimation from the distributed implementation.

There is considerable room for further optimization. First, if we were
pushing for the largest achievable scale, we would use a C++
implementation of the map, reduce and shuffle
operations. Back-of-the-envelope suggestions suggest performance gains
of 100x should be feasible with only minimal memory hierarchy
optimization. Second, we would focus on use of a small number of
many-core machines. Third, we would use a distributed framework such
as HaLoop\footnote{\url{https://code.google.com/p/haloop/}} or MPI, which would
permit simultaneous, asynchronous computation and communication. Fourth, it is known that tuning various parameters that control Hadoop can result in significant performance enhancements \citep{herodotou2011profiling}.
For datasets larger than 100GB (roughly 50x larger than the datasets
considered in this paper), we would want to distribute not just the
latent state, but also the data itself, perhaps using the Hadoop File
System \citep{shvachko}. These advanced distributed implementation
techniques are beyond the scope of this paper.

\cite{williamson2013}, concurrently with and independently of our
previous preliminary work \citep{lovell2013}, investigated a related
parallel MCMC method based on the same auxiliary variable scheme. They
focus on multi-core but single-machine implementations and on
applications to admixtures based on the hierarchical Dirichlet process
(HDP) \citep{teh-etal-2006a}. The compatibility of our transition
operators with a Map-Reduce implementation enables us to analyze
datasets with 100x more dimensions than those from Williamson, even in the
presence of significant inter-machine communication overhead. We also
rely purely on MCMC throughout, based on initialization from our
prior, avoiding the need for heuristic initialization based on
k-means. Combined, our approaches suggest many models based on the DP
may admit principled parallel schemes and scale to significantly
larger problems than are typical in the literature. 

Intuitively, this scheme resembles running "restricted Gibbs" scans over subsets of the clusters, then shuffling clusters amongst subsets, which one might expect to yield slower convergence to equilibrium than full Gibbs scans. Our auxiliary variable representation shows this can be interpreted in terms of exact transition operations for a DPM.

\begin{figure}[t]
  \centering
  \includegraphics[width=.48\textwidth]{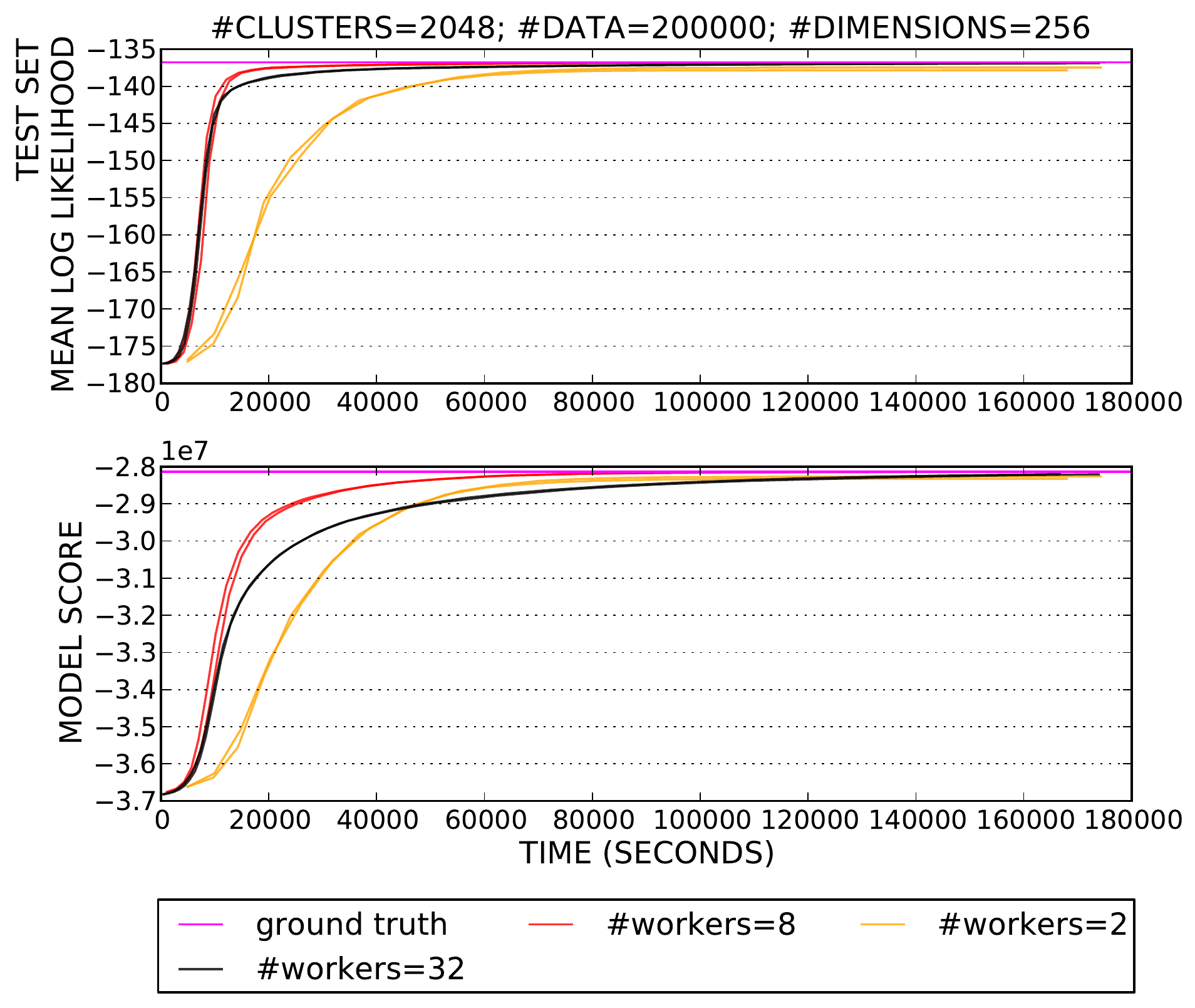}%
  \hfill%
  \includegraphics[width=.48\textwidth]{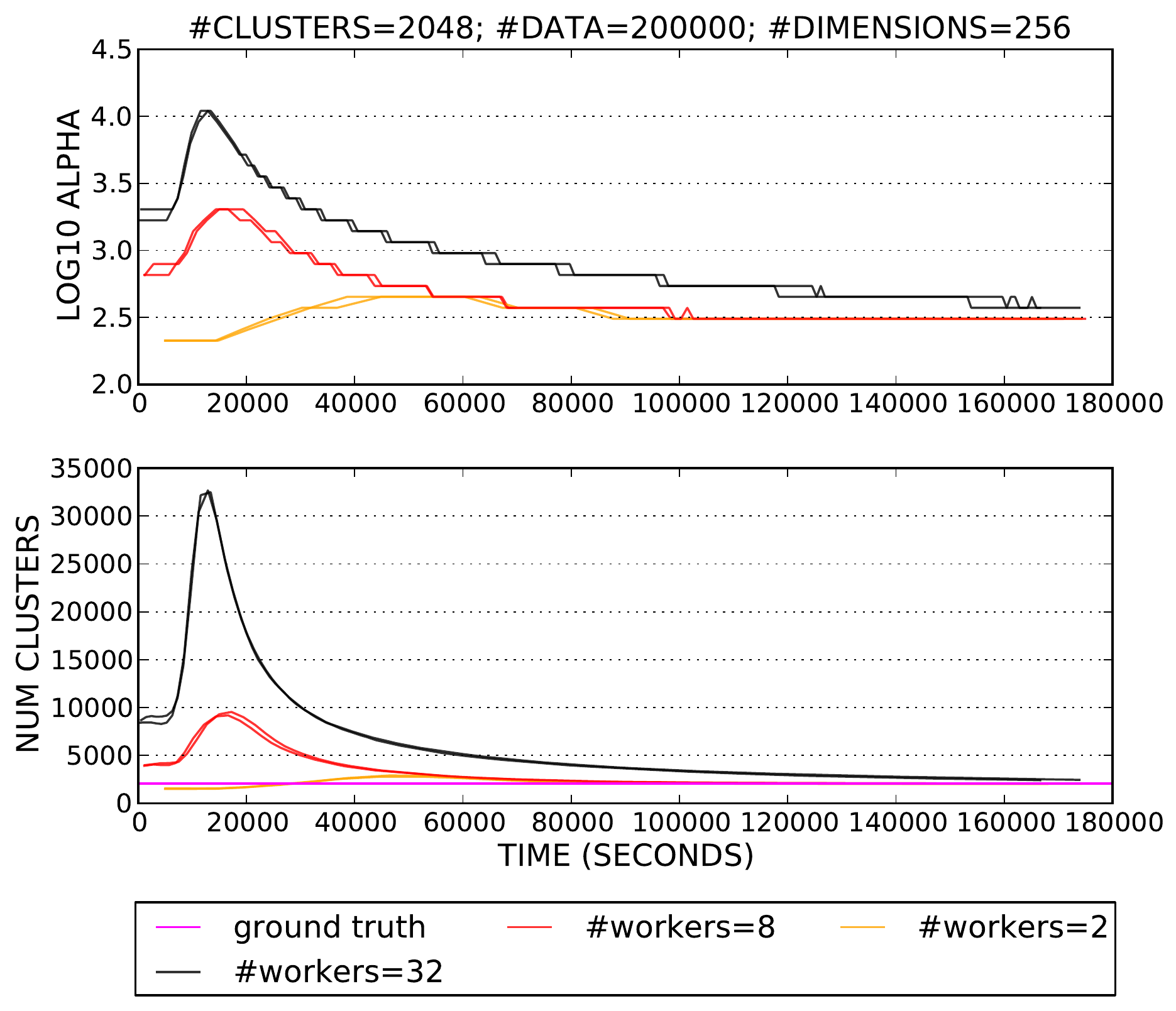}%
  \caption{{\bf Predictive density estimates converge quickly, while latent structure estimates converge more slowly.} (top) Results for our parallel sampler on a synthetic dataset consisting of 2048 clusters and 200,000 datapoints. Each line represents a different number of compute nodes used to parallelize the sampler:  either 2, 8, or 32 nodes were used. The purple line represents ground truth. The parallel samplers perform correctly as determined by eventual convergence to the true likelihood of the test set. Parallel gains are seen up for up to 8 compute nodes, at which point we reach saturation. (bottom) The parallel samplers also eventually convert to the true number of clusters, but at a much slower rate than convergence to the predictive log likelihood. Two runs with different inferential random seeds are shown for each configuration. See the main text for further discussion.}

  \label{fig:convergence}
\end{figure}

\section{Experiments}

We explored the accuracy of our prototype distributed implementation
on several synthetic and real-world datasets. Our synthetic data was drawn from a balanced finite mixture model. Each mixture component $\theta_j$ was parameterized by a set of coin weights drawn from a $\text{Beta}(\beta_d, \beta_d)$ distribution, where $\{\beta_d\}$ is a set of cluster component hyperparameters, one per dimension $d$ of the data. The binary data were Bernoulli draws based on the weight parameters $\{\theta_j\}$ of their respective clusters. Our implementation collapsed out the coin weights and updated each $\beta_d$ during the reduce step using a Griddy Gibbs \citep{ritter} kernel.

\begin{figure}[t]
  \centering
  \includegraphics[width=.49\textwidth]{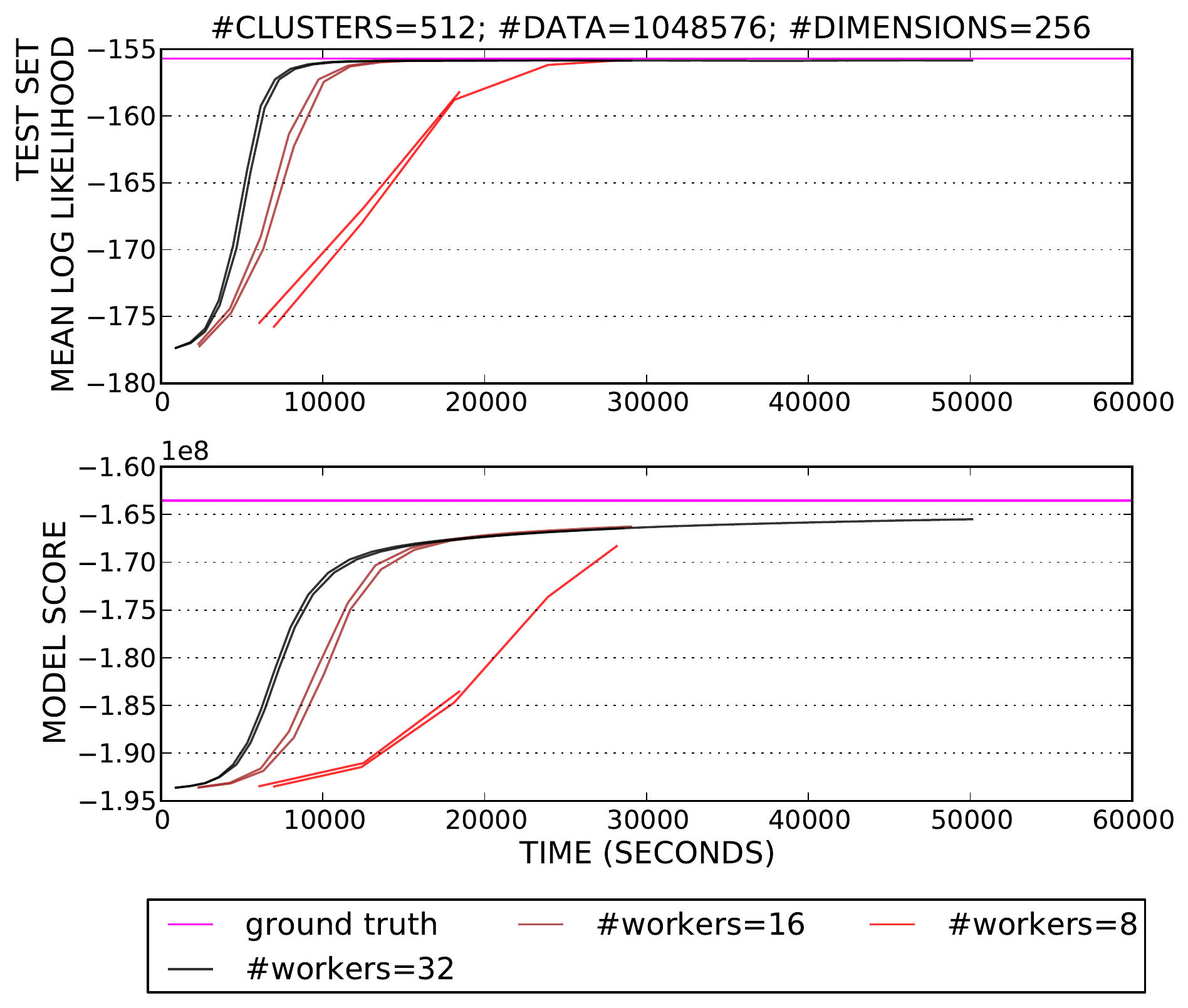}%
  \hfill%
  \includegraphics[width=.49\textwidth]{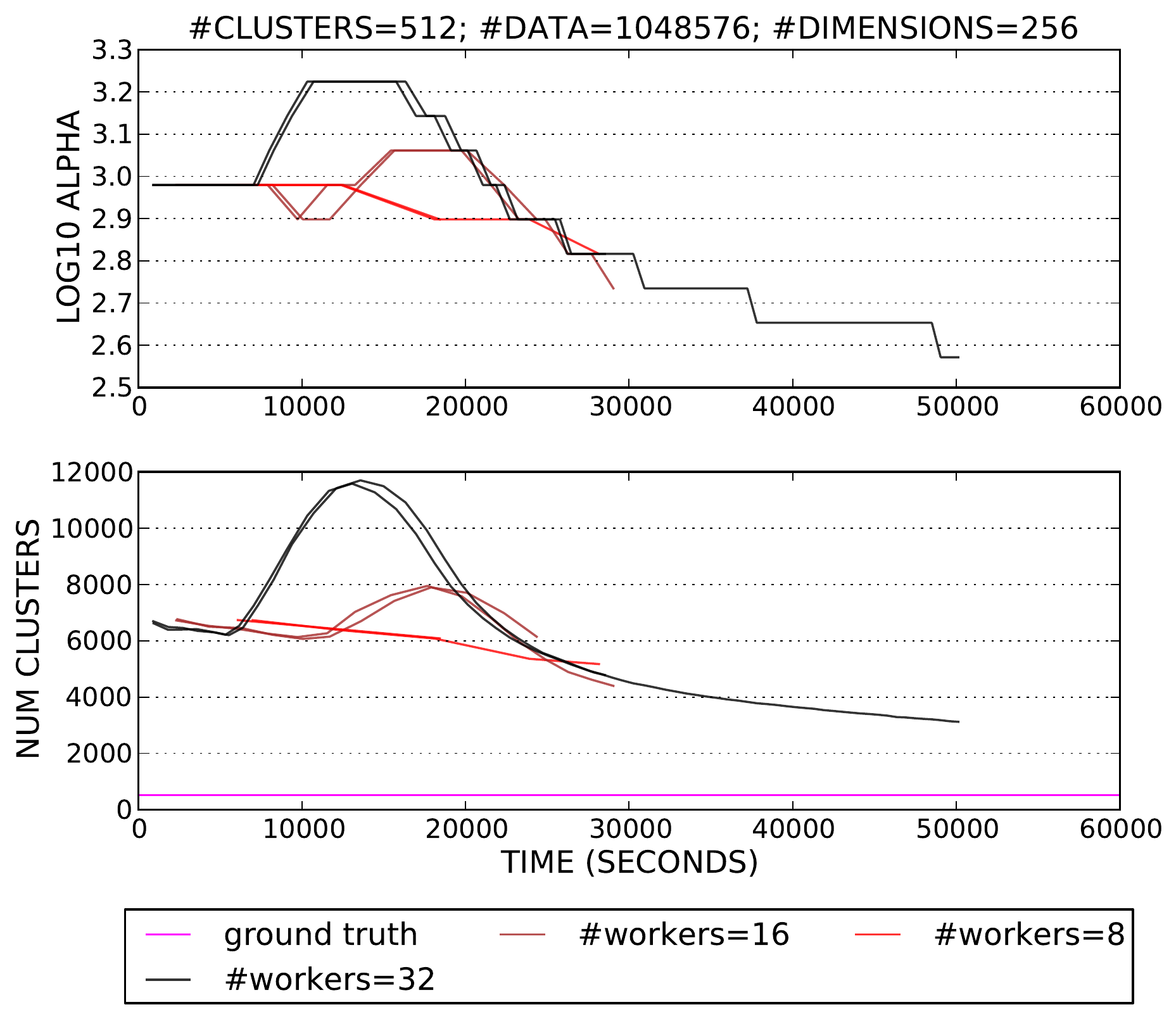}
  \caption{\textbf{Parallel efficiencies for 32 workers can be seen with 1MM rows and 512 clusters.} Consistent with our numerical calculations, larger datasets with more clusters afford more opportunities for parallel gains. At this scale, larger than the one from Figure \ref{fig:convergence}, we see parallel efficiencies up to 32 workers and no slowdown in latent structure convergence.
  }
  \label{fig:parallel_gains_big_learning}
\end{figure}

Figure \ref{fig:predictive_estimates} shows results supporting the accuracy
of our inference scheme as a density estimator given high-dimensional
datasets with large numbers of clusters. We see reliable convergence
to predictive probabilities close to the true entropy of the
generating mixture.

\begin{figure}[t]
  \centering%
  \begin{minipage}[t]{0.48\linewidth}
    \centering%
    \includegraphics[height=2.3in]{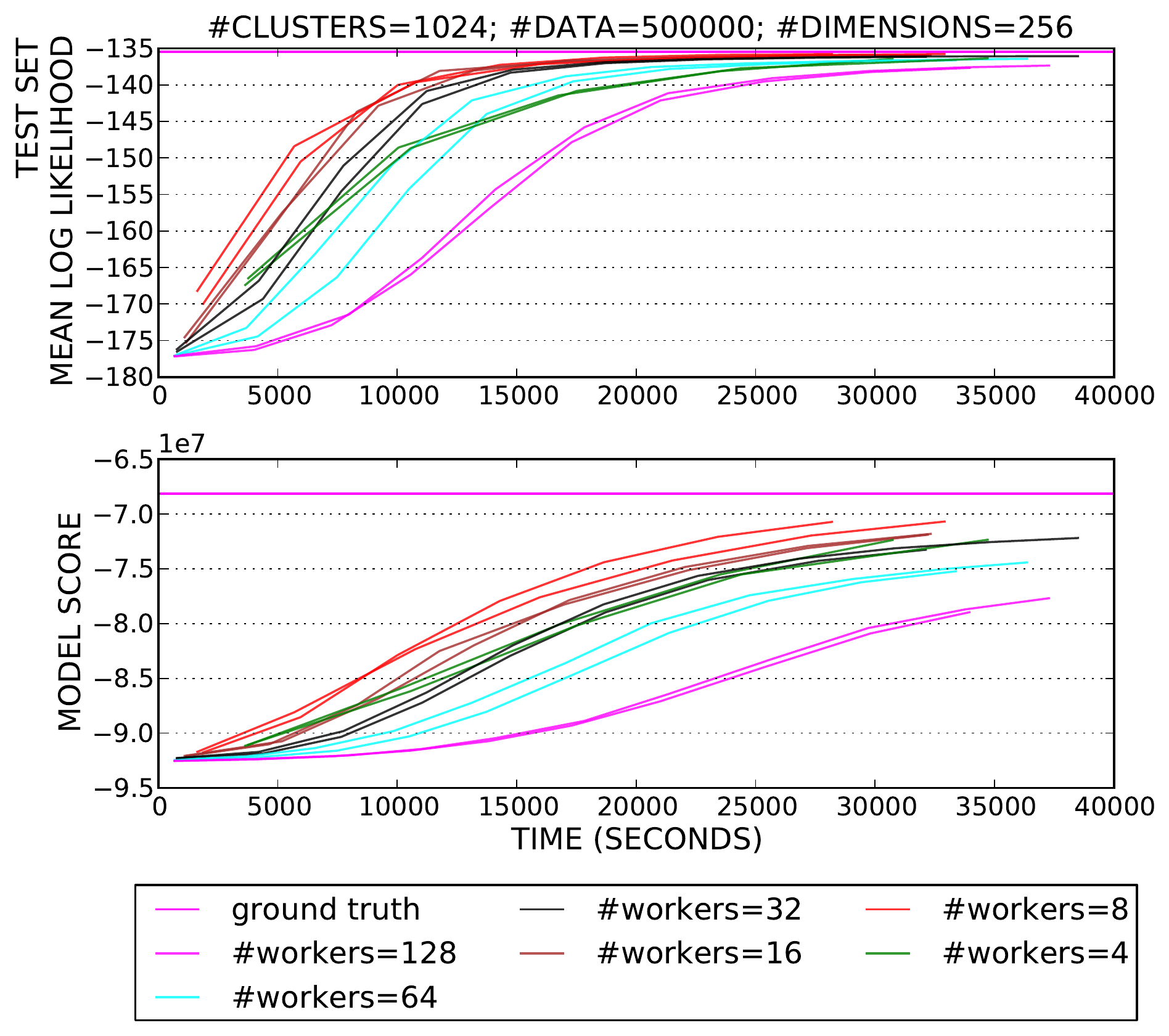}
    \caption{\textbf{Saturation as communication costs and convergence slowdown overwhelm per-iteration parallelism gains.} Results on a 500,000 row problem with 1024 clusters, including 2, 8, 32 and 128 compute nodes (for a max of 64 machines), showing more rapid convergence up to saturation, then slower convergence afterwards.
    }
  \label{fig:parallel_gains_recent}
  \end{minipage}
  \hfill
  \begin{minipage}[t]{0.48\linewidth}
    \centering%
  \includegraphics[height=2.3in]{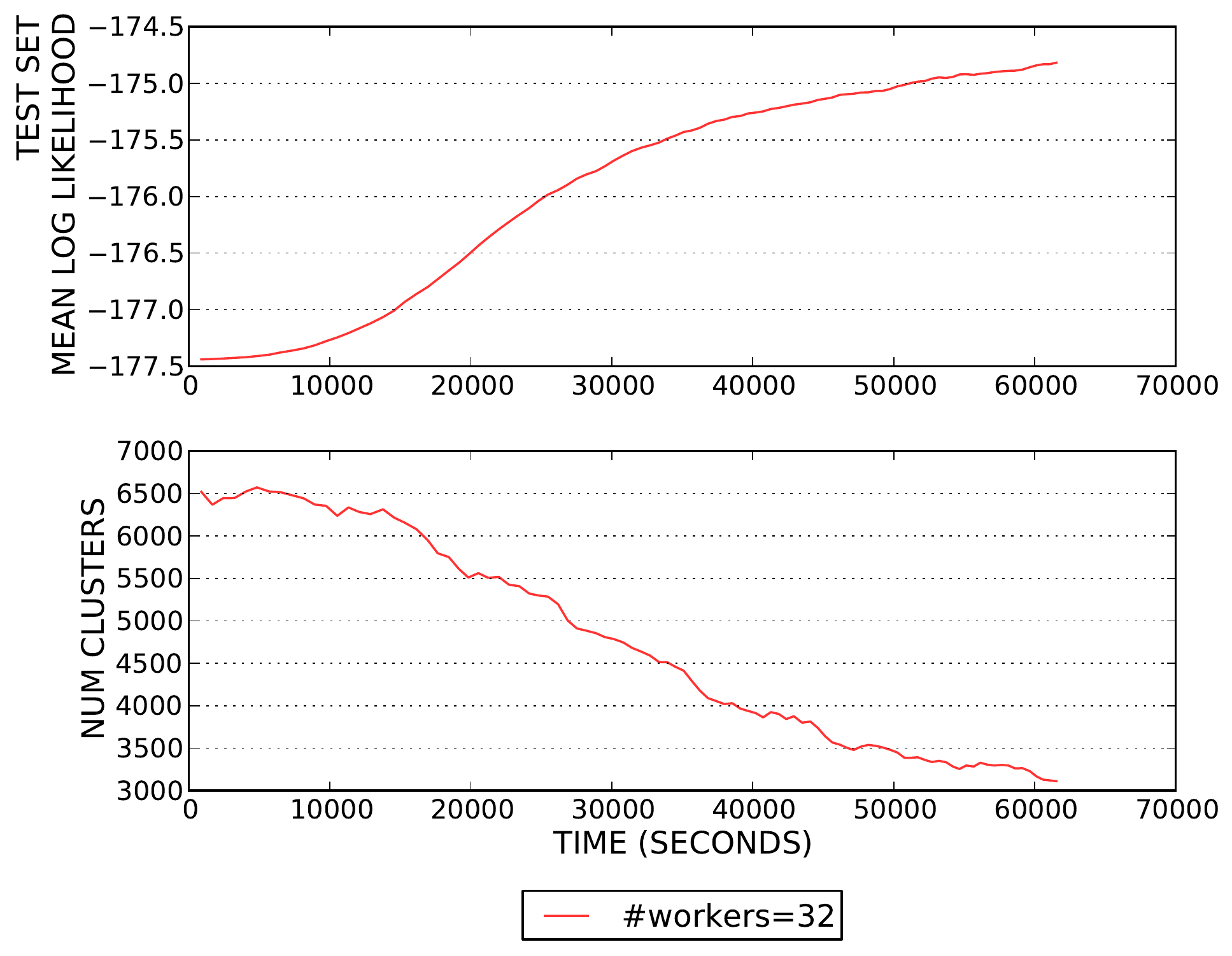}
  \caption {{\bf An illustration on vector quantization of a 1MM
      subset of the Tiny image dataset with 32 workers.}
    Convergence of a representative run in terms of predictive
    accuracy and number of clusters.}
  \label{fig:tinyimages1}
  \end{minipage}
\end{figure}

Figure \ref{fig:convergence} shows the convergence behavior of our
sampler: predictive densities (and joint probabilities) typically
asymptote quickly, while latent structure estimates (such as the
number of clusters, or the concentration parameter) converge far more
slowly. As the Dirichlet process is known to not result in consistent
estimates of number of clusters for finite mixture datasets
\citep{miller-harrison-2013a} (which have no support under the DP
prior), it is perhaps not surprising that predictive likelihood
converges more quickly than estimates of number of
clusters~--~especially given our auxiliary variable representation,
which may encourage representations with multiple predictively
equivalent clusters. It would be interesting to characterize the
regimes where these dynamics occur, and to determine whether they are
also present for approximate parallelization schemes or variational
algorithms based on our auxiliary variable representation. Figure \ref{fig:parallel_gains_recent} shows a typical pattern of efficiencies for parallel computation: for a given problem size, speed increases until some saturation point is reached, after which additional compute nodes slow down the computation. In future work we will explore means of separating the components of this tradeoff due to communication costs, initialization, and any components due to convergence slowdown.

\begin{figure}
  \centering
  \includegraphics[width=\textwidth]{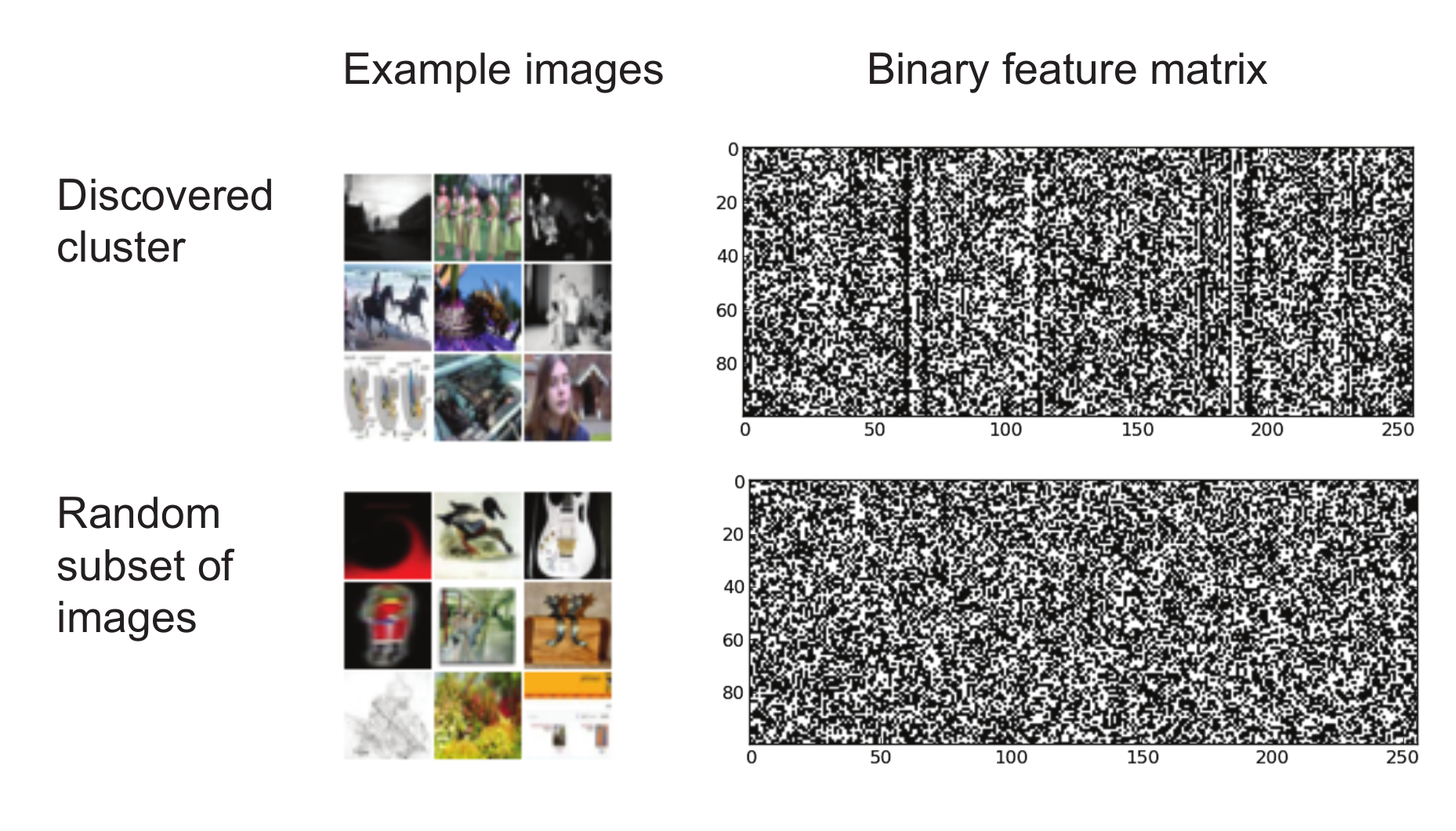}
  \caption{ (left) 9 input images from a
    representative cluster, versus 9 random input images, showing less
    visual coherence. (right) 100 binary feature vectors from a single
    inferred cluster, versus 100 random binary feature vectors from
    the dataset, showing significant compression.}
  \label{fig:tinyimages2}
\end{figure}

Finally, in Figures \ref{fig:tinyimages1} and \ref{fig:tinyimages2},
we show a representative run on a 1MM vector subset of the Tiny Images
\citep{torralba2008} dataset, where we use the Dirichlet process
mixture to perform vector quantization. The input data are binary
features obtained by running a randomized approximation to PCA on
100,000 rows and thresholding the top 256 principal components into
binary values at their component-wise median. After one day of elapsed
time and 32 CPU-days of computation, the sampler is still making
significant progress compressing the data, and has converged to the
vicinity of 3000 clusters. Serial MCMC (not shown) is intractable on
this problem.

\section{Conclusion}
We have introduced an auxiliary variable representation of the
Dirichlet process and applied it to mixture models, where it yields
superclusters that cluster the clusters. We have shown how this
representation enables an exact parallelization of the standard MCMC
schemes for inference, where the DP learns how to parallelize itself,
despite the lack of conditional independencies in the traditional form
of the model. We have also shown that this representation naturally
meshes with the Map-Reduce approach to distributed computation on a
large compute cluster, and developed a prototype distributed
implementation atop Amazon's Elastic Compute Cloud, tested on over 50
machines and 100 cores. We have explored its performance on synthetic
and real-world density estimation problems, including runs on over 1MM
row, 256-dimensional data sets.

These results point to a potential path forward for ``big data''
applications of Bayesian statistics for models that otherwise lack
apparent conditional independencies. We suspect searching for
auxiliary variable representations that induce independencies may lead
to new ways to scale up a range of nonparametric Bayesian models, and
may perhaps also lead to further correspondences with established
distributed computing paradigms for MCMC inference. We hope our
results present a useful step in this direction.

\FloatBarrier

\subsection*{Acknowledgments}

This work was partially supported by DARPA XDATA contract FA8750-12-C-0315. RPA was funded in part by DARPA Young Faculty Award N66001-12-1-4219.  VKM was partially supported by gifts from Google and Analog Devices.

\bibliographystyle{unsrtnat}
\bibliography{draft}

\end{document}